\documentclass{article}

\usepackage[dblblindworkshop, final]{neurips_2026}
\workshoptitle{AI for Stochastic Dynamics}
\usepackage[T1]{fontenc}
\usepackage{graphicx}
\usepackage{amsmath,amssymb}
\usepackage{booktabs}
\usepackage{float}
\usepackage{placeins}
\usepackage[colorlinks=true,linkcolor=blue,citecolor=blue,urlcolor=blue]{hyperref}
\usepackage{adjustbox}

\setcitestyle{round,aysep={}}
\newcommand{\mse}[2]{$#1\!\pm\!#2$}
\newcommand{\bestmse}[2]{$\mathbf{#1\!\pm\!#2}$}
\newcommand{\secondmse}[2]{$\underline{#1\!\pm\!#2}$}

\newcommand{\SupportCal}{\textsc{SupportCal}}

\title{SupportCal: Label-Free Calibration of Post-Trained LLMs via Reference Support and Corroboration}

\author{
\textbf{Linhan Luo}$^{1}$ \quad
\textbf{Lequan Lin}$^{1}$ \quad
\textbf{Dai Shi}$^{2}$ \\[2pt]
\textbf{Feng Chen}$^{3}$ \quad
\textbf{Jos\'e Miguel Hern\'andez-Lobato}$^{2}$ \quad
\textbf{Junbin Gao}$^{1}$
}

\date{}

\begin{document}

\maketitle

\begingroup
\renewcommand{\thefootnote}{}
\footnotetext{\noindent
$^{1}$ The University of Sydney \quad
$^{2}$ University of Cambridge \quad
$^{3}$ The University of Adelaide\\
Correspondence: \texttt{lluo2926@uni.sydney.edu.au}}
\addtocounter{footnote}{-1}
\endgroup

\begin{abstract}
Post-training often improves task performance but can degrade confidence calibration, leaving post-trained language models (PoLMs) more overconfident than their corresponding pretrained language models (PLMs). 
Because task-specific labeled calibration data can be costly or unavailable, the corresponding PLM provides a natural label-free reference for post-hoc calibration. 
Prior agreement-gated PLM-referenced calibration fits a scalar temperature using only examples on which the PoLM and its PLM reference agree, excluding disagreement examples because direct alignment can drive the fitted temperature excessively high and induce under-confidence.
We revisit this binary treatment. A controlled reintroduction diagnostic reveals a non-monotonic aggregate effect: admitting a moderate fraction of disagreement examples can improve calibration, whereas the benefit diminishes as unit-weight inclusion approaches the full disagreement set.
We introduce \SupportCal{}, a label-free post-hoc method that retains agreement examples at unit weight and assigns disagreement examples continuous weights based on the own-base PLM's relative support and corroboration from pretrained references selected from a size-compatible candidate pool. 
We further characterize when the resulting weighted objective admits a finite optimal temperature.  Across MedMCQA and MathQA, \SupportCal{} yields lower mean ECE than the agreement-only baseline for nearly all evaluated target-model configurations;
supplementary TweetEval Sentiment results show the same pattern on a fixed-label classification task.
\end{abstract}

\section{Introduction}
\label{sec:intro}

\begin{figure}[t]
\centering
\includegraphics[width=\linewidth]{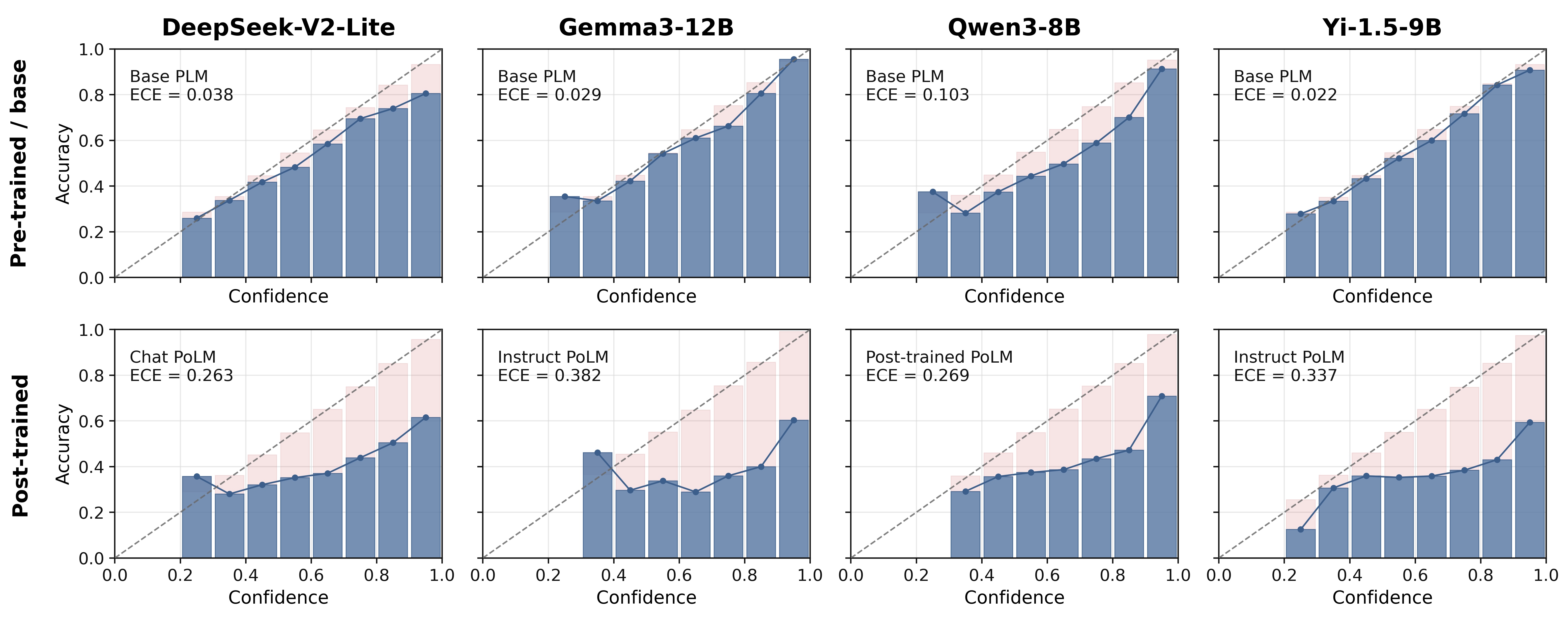}
\caption{Reliability diagrams on MedMCQA for representative own-base PLM/PoLM pairs. Base PLMs are generally better calibrated than their post-trained counterparts. ECE annotations use the $[0,1]$ scale (e.g., $0.038=3.8$ pp).}
\label{fig:reliability}
\end{figure}

A reliable model should not only answer correctly but also report confidence that matches its empirical frequency of correctness \citep{guo2017calibration}. 
Post-training, including instruction tuning and reinforcement learning from human feedback, can improve downstream performance and instruction-following utility while degrading confidence calibration relative to the corresponding pretrained language model (PLM) \citep{ouyang2022instruct, zhu2023calibration}. 
We refer to that pretrained counterpart as the target's \emph{own-base PLM}; for example, an instruction-tuned Llama-3-8B target is paired with the corresponding pretrained Llama-3-8B checkpoint. 
Across the representative pairs in Figure~\ref{fig:reliability} and in prior studies, own-base PLMs often remain better calibrated than their post-trained models \citep{luo2025daca, tan2026basecal}. 
Because task-specific labeled calibration data can be costly or unavailable, this better-calibrated model provides a natural label-free reference for post-hoc calibration.

Prior agreement-gated PLM-referenced calibration fits a single post-hoc temperature by KL alignment to a PLM reference \citep{luo2025daca}. On disagreement examples, however, the reference PLM's highest-probability option differs from the PoLM's prediction. Directly aligning these distributions can drive the fitted temperature excessively high and induce under-confidence. 
Prior work therefore assigns zero weight to every disagreement example and fits the temperature using agreement examples only. This safeguard addresses a genuine failure mode, but it also discards the entire disagreement region.

This raises a natural question:
\emph{Must every disagreement example be discarded?}
Disagreement is common in our evaluation, and the controlled diagnostic in Section~\ref{sec:motivation} reveals a non-monotonic aggregate effect: admitting a moderate fraction of disagreement examples can improve calibration, whereas the benefit diminishes as unit-weight inclusion approaches the full disagreement set.
This pattern shows that binary exclusion is too coarse and motivates controlling how strongly each disagreement example influences temperature fitting.

We introduce \SupportCal{}, an agreement-preserving continuous weighting method. Agreement examples retain unit weight. Disagreement examples receive graded label-free weights that combine the own-base PLM's relative support for the PoLM prediction with corroboration from pretrained references selected from a size-compatible candidate pool. A single temperature is then fitted by weighted KL alignment to the target's own-base PLM, which remains the sole KL teacher.

\noindent\textbf{Contributions.}
\textbf{(1) Diagnosis.} A controlled disagreement-reintroduction experiment reveals a non-monotonic aggregate effect: moderate reintroduction can improve calibration, whereas the benefit diminishes as unit-weight inclusion approaches the full disagreement set.
\textbf{(2) Method and analysis.} \SupportCal{} preserves agreement examples at unit weight and assigns disagreement examples graded weights using own-base relative support and cross-reference corroboration. A boundary characterization gives the condition under which the weighted KL objective has a finite optimal temperature.
\textbf{(3) Empirical validation.} Across MedMCQA and MathQA, \SupportCal{} yields lower mean ECE than the agreement-only baseline for nearly all evaluated target-model configurations. Supplementary TweetEval Sentiment results (Appendix~\ref{app:tweeteval}) show the same pattern beyond multiple-choice question answering.

\section{Background and Related Work}
\label{sec:background}

\subsection{Calibration preliminaries}
\label{sec:preliminaries}

Consider a $K$-class prediction problem, including multiple-choice question answering and fixed-label classification. 
For example $n$, let $y_n\in\{1,\ldots,K\}$ be the ground-truth label and
$z_n=(z_{n,1},\ldots,z_{n,K})$ be the model logits. We define
$p_n=\operatorname{softmax}(z_n)$, prediction
$\hat y_n=\arg\max_k p_{n,k}$, and confidence
$\gamma_n=\max_k p_{n,k}$.

Post-hoc temperature scaling with $\tau>0$ gives
$q_n(\tau)=\operatorname{softmax}(z_n/\tau)$ and calibrated confidence
$\gamma_n(\tau)=\max_k q_{n,k}(\tau)$.
Because division by a positive scalar preserves the ordering of the logits, temperature scaling does not change $\hat y_n$. At the population level, a model is perfectly calibrated when
\[
\Pr(\hat Y=Y\mid \Gamma=\gamma)=\gamma
\]
for every confidence level $\gamma$ in the support of $\Gamma$ \citep{guo2017calibration}.

We measure empirical miscalibration using Expected Calibration Error (ECE)
\citep{naeini2015bbq, guo2017calibration}. Let $\{\mathcal{B}_b\}_{b=1}^{B}$ be $B=10$ equal-width confidence bins. Then
\begin{equation}
\mathrm{ECE}
=
\sum_{b=1}^{B}
\frac{|\mathcal{B}_b|}{N}
\left|
\mathrm{acc}(\mathcal{B}_b)
-
\mathrm{conf}(\mathcal{B}_b)
\right|,
\label{eq:ece}
\end{equation}
where $\mathrm{acc}(\mathcal{B}_b)$ and $\mathrm{conf}(\mathcal{B}_b)$ are the empirical accuracy and mean confidence in bin $b$, respectively. We report ECE in percentage points (pp) and use the same ten-bin protocol for all methods.

\subsection{LLM confidence estimation and calibration}

Work on LLM confidence spans confidence elicitation, uncertainty estimation, and calibration \citep{geng2024survey}. Calibration can change substantially across the training pipeline: \citet{zhu2023calibration} find that alignment training often increases overconfidence even when task performance improves.
Earlier work shows that pretrained Transformer encoders can be reasonably calibrated in-domain \citep{desai2020calibration}, while large language models can contain useful self-assessment signals about their own correctness \citep{kadavath2022know}. 
More recent PLM-referenced studies further observe that corresponding base models often remain better calibrated than their post-trained counterparts \citep{luo2025daca, tan2026basecal}. Together, these results motivate using pretrained-model information as an external calibration signal rather than relying exclusively on the PoLM's own confidence.

Existing approaches obtain or improve confidence through several mechanisms.
\emph{Supervised post-hoc methods}, such as temperature scaling, fit a calibration map using correctness-labeled data \citep{guo2017calibration}.
\emph{Direct self-assessment methods} elicit $P(\mathrm{True})$ or verbalized confidence from the model \citep{kadavath2022know, lin2022teaching, tian2023justask, xiong2024uncertainty}.
\emph{Sampling-based methods} estimate uncertainty from consistency across multiple generations or from entropy over semantically equivalent answers \citep{kuhn2023semantic, farquhar2024semanticentropy}. These methods can capture uncertainty in free-form generation but require multiple inference calls. 
\emph{Prompt-ensemble methods}, such as CAPE, instead aggregate predictions across answer-option permutations or related prompt variants \citep{jiang2023cape}. 
Finally, \emph{training-based methods} teach models to express calibrated confidence through supervised fine-tuning or reinforcement learning \citep{kapoor2024calibrationtuning, xu2024sayself}.

Recent methods also exploit signals outside the PoLM's original confidence. BaseCal uses the corresponding base model either to re-evaluate PoLM generations or to train a lightweight hidden-state projection without correctness labels \citep{tan2026basecal}. 
CaliDist instead constructs an instance-specific robustness signal from prediction changes under semantic distractors and fits a task-specific scaling map on a held-out labeled validation set \citep{jawad2026calidist}. 
Our setting instead uses pretrained references to construct sample weights for a single post-hoc temperature fitted against the target's own-base PLM.

\subsection{Closest prior work: PLM-referenced calibration}

DACA \citep{luo2025daca} is the closest prior work to our setting. It performs label-free temperature fitting by aligning a PoLM to a PLM reference while excluding disagreement examples from the objective to avoid disagreement-induced over-softening.

\SupportCal{} instead treats disagreement as a graded weighting problem. 
Rather than discarding all disagreement examples, it assigns each example a continuous calibration weight according to two complementary signals: the own-base PLM's relative support for the PoLM prediction and corroborating support from selected pretrained references. 
Agreement examples remain fully weighted, while disagreement examples contribute in proportion to the strength of this combined support.
This design allows information from disagreement examples to be recovered without treating all of them as equally reliable. The resulting weights are used in a weighted KL objective to fit a single scalar temperature against the target's own-base PLM. 

\section{Motivating Diagnosis: Reintroducing Disagreement}
\label{sec:motivation}

An agreement-only objective excludes all calibration examples on which the target PoLM and its own-base PLM predict different options. We examine the aggregate effect of this binary treatment by progressively reintroducing disagreement examples during temperature fitting.

Disagreement is substantial in the broader diagnostic model pool: the median own-base PLM/PoLM disagreement rate is 28.9\% on MedMCQA and 46.7\% on MathQA. Per-target disagreement rates for this broader pool are reported in Appendix~\ref{app:disagreement}.

\begin{figure}[t]
\centering
\includegraphics[width=\linewidth]
{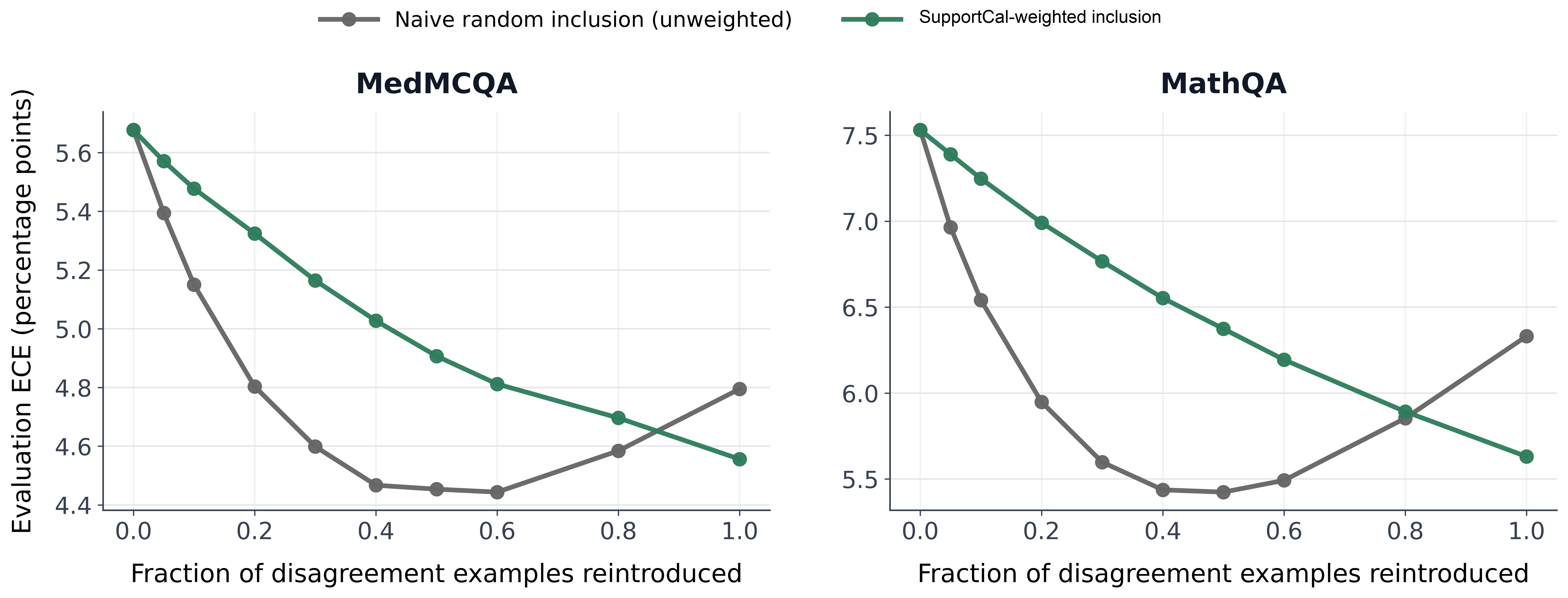}
\caption{Controlled disagreement reintroduction on MedMCQA and MathQA.
The gray curve assigns unit weight to reintroduced disagreement examples,
whereas the green curve applies \SupportCal{} weights.
Each point averages over target models and five data-split seeds; lower ECE is better.}
\label{fig:reintroduction}
\end{figure}

Figure~\ref{fig:reintroduction} reveals a non-monotonic effect under unit-weight inclusion. On both datasets, ECE initially decreases, reaches its best region at a moderate inclusion fraction (roughly $0.4$--$0.6$), and then rises as full inclusion is approached. The non-monotonic trend motivates sample-specific weighting within the disagreement region.

When the same examples are scaled by \SupportCal{} weights, ECE continues to improve through the full-inclusion endpoint. The diagnostic therefore motivates preserving agreement examples while assigning sample-specific influence within the disagreement region.

\section{\SupportCal{}}
\label{sec:method}

\begin{figure}[t]
    \centering
    \includegraphics[width=\textwidth]{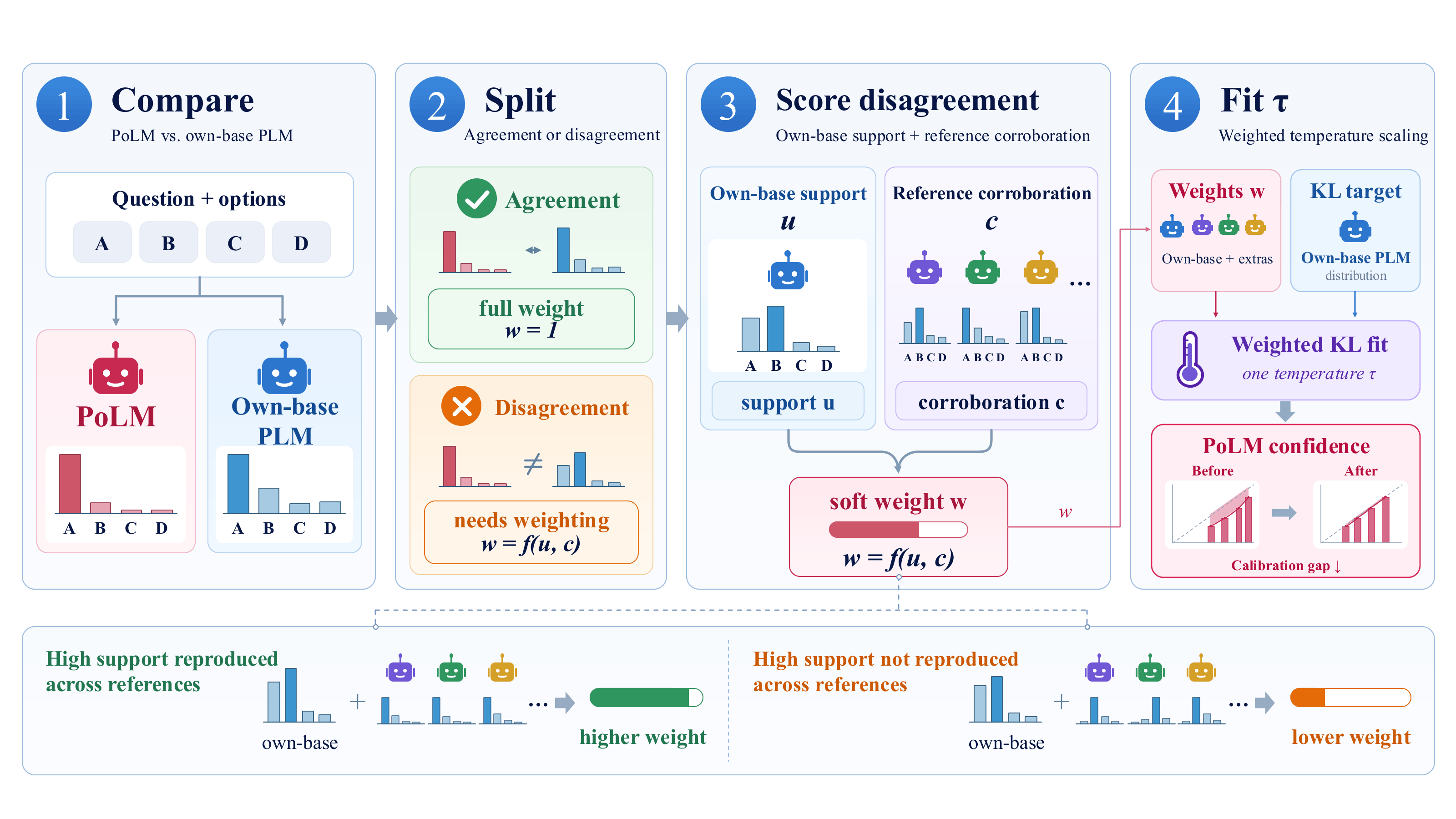}
    \caption{
    Overview of \SupportCal. Agreement examples retain unit weight; disagreement examples are weighted by own-base relative support and cross-reference corroboration before fitting a single temperature against the own-base PLM.
    }
    \label{fig:supportcal_framework}
\end{figure}

\SupportCal{} preserves agreement examples and assigns support-aware weights to disagreement examples before fitting a single temperature. Figure~\ref{fig:supportcal_framework} summarizes the overall pipeline; the following subsections define each component.

\subsection{Relative support on disagreement examples}

Using the notation of Section~\ref{sec:preliminaries}, the target PoLM produces logits $z_n$ and prediction $\hat y_n$ for calibration example $n=1,\ldots,N$. 
Its own-base PLM, the corresponding pretrained checkpoint from which the PoLM is derived, produces distribution $p_n^{(o)}=\operatorname{softmax}(v_n^{(o)})$ and prediction $\hat y_n^{(o)}=\arg\max_k p_{n,k}^{(o)}$. 
All model distributions are evaluated over the same $K$ answer options.

We partition the calibration split into the agreement and disagreement regions
\[
A=\{n:\hat y_n^{(o)}=\hat y_n\},
\qquad
D=\{n:\hat y_n^{(o)}\neq\hat y_n\}.
\]
\SupportCal{} retains unit weight on $A$ and constructs continuous, label-free weights for examples in $D$.

For any pretrained reference $r$ with distribution $p_n^{(r)}=\operatorname{softmax}(v_n^{(r)})$, we define its \emph{relative support} for the PoLM prediction as
\begin{equation}
u_n^{(r)}
=
\frac{p_{n,\hat y_n}^{(r)}}
{\max_k p_{n,k}^{(r)}}
\in(0,1].
\label{eq:relsupport}
\end{equation}

This ratio measures how strongly reference $r$ supports the PoLM-selected option relative to its own most likely option. On a disagreement example, a small own-base score $u_n^{(o)}$ indicates a sharp contradiction: the own-base PLM assigns substantially less probability to the PoLM-selected option than to its own preferred option. Down-weighting such examples limits the strong alignment pressure that could otherwise drive the fitted temperature upward and over-soften the PoLM distribution.

\paragraph{Motivation for cross-reference corroboration.}
A large $u_n^{(o)}$, however, is not always equally informative. It may reflect a genuine near-tie, where the own-base PLM still provides substantial support for the PoLM-selected option, but it can also arise from a diffuse or weak own-base distribution. 
We therefore use additional pretrained references as a corroboration signal: a disagreement example receives stronger influence when support for the PoLM-selected option is reproduced across the selected references, and is attenuated when that support is not corroborated. Appendix~\ref{app:relative_only_failure} provides representative comparisons illustrating the benefit of this additional signal.

\subsection{Cross-reference corroboration and selection}

For each target, let $R$ denote the candidate pool of pretrained references from the same predefined parameter-size bucket, excluding the target's own-base PLM. For a fixed $M$, we consider every eligible combination
$S\subseteq R$ with $|S|=M$.

For a candidate combination $S$, we first define its sample-level corroboration as the mean relative support across the references in $S$:
\begin{equation}
c_n(S)
=
\frac{1}{M}
\sum_{r\in S} u_n^{(r)}.
\label{eq:corroboration}
\end{equation}

We then score the combination by its average corroborating support over the own-base disagreement region:
\begin{equation}
\mathrm{DS}(S)
=
\frac{1}{|D|}
\sum_{n\in D} c_n(S).
\label{eq:ds}
\end{equation}

For the fixed $M$, we enumerate all eligible size-$M$ combinations and select
\begin{equation}
S^\star
\in
\arg\max_{\substack{S\subseteq R\\|S|=M}}
\mathrm{DS}(S).
\label{eq:selector}
\end{equation}

The selected combination therefore consists of the extra pretrained references that collectively provide the strongest average support for the PoLM prediction over the calibration disagreement region. All combination scores are computed using only model outputs from the unlabeled calibration split.

Finally, we write $c_n^\star=c_n(S^\star)$ for the sample-level corroboration supplied by the selected reference combination.

\subsection{Support-aware weighting and temperature fitting}

The final sample weight is
\begin{equation}
w_n
=
\begin{cases}
1, & n\in A,\\[2pt]
u_n^{(o)}c_n^\star, & n\in D.
\end{cases}
\label{eq:weights}
\end{equation}
The multiplicative form acts as a soft conjunction: a disagreement example receives a large weight only when both the own-base PLM and the selected references provide strong relative support for the PoLM prediction. Agreement examples retain full influence.

\SupportCal{} fits a single scalar temperature using the weighted KL objective
\begin{equation}
\hat\tau
=
\arg\min_{\tau>0}
\frac{
\sum_n
w_n
D_{\mathrm{KL}}
\!\left(
p_n^{(o)}
\,\middle\|\,
\operatorname{softmax}(z_n/\tau)
\right)}
{\sum_n w_n}.
\label{eq:tempobj}
\end{equation}
Cross-reference corroboration determines how strongly each example influences the fit, while the own-base distribution defines the calibration target.

\subsection{Finite-temperature characterization}
\label{sec:boundary}

Let $\beta=1/\tau$,
$q_{n,\beta}=\operatorname{softmax}(\beta z_n)$, and
\[
\ell_n(\beta)
=
D_{\mathrm{KL}}
\!\left(
p_n^{(o)}
\,\middle\|\,
q_{n,\beta}
\right),
\qquad
L(\beta)
=
\frac{1}{W}
\sum_n w_n\ell_n(\beta),
\qquad
W=\sum_n w_n.
\]
Define the unweighted logit mean
$\bar z_n=\frac{1}{K}\sum_k z_{n,k}$, the per-example
\emph{boundary margin}
\[
m_n
=
\mathbb{E}_{p_n^{(o)}}[z_n]
-
\bar z_n,
\]
and the weighted boundary margin
\[
M_w
=
\frac{1}{W}
\sum_n w_n m_n.
\]

Because finite-logit softmax distributions assign positive probability to every class, Eq.~\eqref{eq:weights} gives $w_n>0$ for every example and hence $W>0$ for any nonempty calibration split. 
If at least one target-logit vector $z_n$ is nonconstant, as in all evaluated configurations, then $L(\beta)$ is strictly convex on $[0,\infty)$.

\textbf{Proposition 1 (finite-temperature characterization).}
Under the condition above,
\[
L'(0)=-M_w.
\]
Consequently, the original temperature objective admits a unique finite minimizer if and only if $M_w>0$. When $M_w>0$, 
there is a unique $\beta^\star\in(0,\infty)$, equivalently a unique finite temperature
$\tau^\star=1/\beta^\star$. 
When $M_w\leq0$, the continuous extension is minimized at $\beta^\star=0$, corresponding to $\tau\to\infty$ in the original temperature domain. 
The proof is given in
Appendix~\ref{app:boundary}.

\textbf{Sample-level interpretation.}
Since $\ell_n'(0)=-m_n$, examples with $m_n<0$ contribute toward the $\beta=0$ boundary, 
whereas examples with $m_n>0$ contribute in the finite-temperature direction. 
The margin $m_n$ therefore gives each example's signed contribution to the boundary behavior, and \SupportCal{} controls this contribution through $w_n$. 
Empirically, negative-margin disagreement examples receive lower mean and median weights than positive-margin examples across all three datasets (Appendix~\ref{app:boundary}).

\section{Experiments}
\label{sec:experiments}

\subsection{Datasets, models, and protocol}

We evaluate \SupportCal{} primarily on two multiple-choice question-answering benchmarks. \textbf{MedMCQA} \citep{pal2022medmcqa} contains medical entrance-exam questions; we evaluate 12 post-trained target models spanning the Gemma 3, Llama 3, Mistral, Qwen2.5, Qwen3, and Yi-1.5 families 
\citep{gemmateam2025gemma3, grattafiori2024llama3, jiang2023mistral7b, yang2024qwen25, yang2025qwen3, young2024yi}.
We report \SupportCal{} with $M\in\{2,4\}$. 
\textbf{MathQA} \citep{amini2019mathqa} contains math word problems;
we evaluate seven target models from the Gemma 3, Llama 3, Qwen2.5, and Yi-1.5 families, including Qwen2.5-Math
\citep{yang2024qwen25math}, and report $M\in\{1,2\}$.

Every target is a post-trained model paired with its corresponding pretrained base checkpoint. 
For each target, the candidate reference pool $R$ contains pretrained models from the same predefined parameter-size bucket, excluding the target's own-base PLM. 
For each fixed $M$, we enumerate all eligible size-$M$ combinations in $R$ and select the combination with the highest Disagreement-Support score. 
The exact bucket membership is given in Appendix~\ref{app:setup}.

Supplementary evaluation on TweetEval Sentiment \citep{barbieri2020tweeteval}, a non-QA fixed-label classification task,
is reported in Appendix~\ref{app:tweeteval}.

For each target model and each of five random seeds, we split examples 30\%/70\% into a calibration split and an evaluation split. 
The calibration split is used for relative-support computation, reference-combination selection, and temperature fitting; the evaluation split is used only to report calibration metrics. 
A single scalar temperature is fitted per seed and applied to the corresponding evaluation split. 
We report mean ECE and standard error across the five seeds separately for each target model.

\subsection{Metrics and baselines}
\label{sec:baselines}

We report Expected Calibration Error (ECE; Section~\ref{sec:preliminaries}) using the same ten-bin protocol for all methods. 
The comparison includes the following methods. 
\textbf{Vanilla} uses the uncalibrated PoLM confidence.
\textbf{CAPE} \citep{jiang2023cape} is a label-free prompt-ensemble baseline; in our implementation, predictions are aggregated across cyclic permutations of the answer-option ordering.
\textbf{Elicitation} \citep{xiong2024uncertainty} prompts the PoLM to generate a predicted answer and verbalized confidence in a structured response.
\textbf{Elicitation-Ensemble} uses the same elicitation procedure; in our implementation, five generations are sampled and their parsed confidence estimates are aggregated.
\textbf{DACA} \citep{luo2025daca} is the agreement-only PLM-referenced baseline, instantiated here using each target's own-base PLM.
\textbf{\SupportCal{} (ours)} assigns continuous support-aware weights and is reported at both tested reference-combination sizes $M$.

\section{Results}
\label{sec:results}

\textbf{Consistent gains over agreement-only calibration.}
On MedMCQA (Table~\ref{tab:medmcqa}), \SupportCal{} yields lower mean ECE than DACA on all 12 target models at both tested values of $M$. 
The macro-average ECE decreases from 6.65 pp for DACA to 5.07 pp for $M=2$ and 5.11 pp for $M=4$, corresponding to reductions of 1.59 and 1.54 pp.

On MathQA (Table~\ref{tab:mathqa}), $M=1$ yields lower mean ECE than DACA on all seven target models, while $M=2$ does so on six of seven. 
The macro-average ECE decreases from 7.79 pp to 5.70 and 5.71 pp, respectively. The only exception is Qwen2.5-14B at $M=2$, where ECE is 0.03 pp higher than DACA.

\begin{table}[H]
\centering
\footnotesize
\caption{
MedMCQA ECE (pp; mean $\pm$ SE over five seeds).
Bold and underline denote the best and second-best methods per target,
respectively, based on unrounded values. The final row reports the model-level macro-average.
}
\label{tab:medmcqa}

\setlength{\tabcolsep}{1.8pt}
\renewcommand{\arraystretch}{1.08}

\begin{adjustbox}{center,max width=0.96\linewidth}
\begin{tabular}{@{}lccccccc@{}}
\toprule
Target PoLM
& Vanilla
& CAPE
& Elicit.
& Elic.-Ens.
& DACA
& \SupportCal{} (2)
& \SupportCal{} (4) \\
\midrule

Gemma3-12B
& \mse{37.965}{0.053}
& \mse{21.310}{0.133}
& \mse{42.925}{0.133}
& \mse{35.133}{0.158}
& \mse{5.352}{0.057}
& \bestmse{4.522}{0.087}
& \secondmse{4.556}{0.095} \\

Gemma3-27B
& \mse{30.085}{0.044}
& \mse{13.691}{0.205}
& \mse{32.856}{0.140}
& \mse{27.682}{0.139}
& \mse{3.901}{0.255}
& \secondmse{3.082}{0.103}
& \bestmse{3.051}{0.107} \\

Gemma3-4B
& \mse{50.407}{0.156}
& \mse{28.622}{0.122}
& \mse{53.553}{0.297}
& \mse{43.355}{0.339}
& \mse{9.238}{0.784}
& \bestmse{2.883}{0.249}
& \secondmse{2.997}{0.224} \\

Llama-3-8B
& \mse{23.146}{0.142}
& \mse{12.948}{0.108}
& \mse{28.483}{0.147}
& \mse{19.736}{0.135}
& \mse{2.988}{0.087}
& \secondmse{2.783}{0.098}
& \bestmse{2.728}{0.124} \\

Mistral-7B-Instruct
& \mse{37.212}{0.062}
& \mse{20.671}{0.043}
& \mse{52.873}{0.239}
& \mse{42.229}{0.281}
& \mse{6.161}{0.188}
& \bestmse{5.049}{0.130}
& \secondmse{5.096}{0.132} \\

Qwen2.5-14B
& \mse{27.337}{0.119}
& \mse{17.180}{0.236}
& \mse{26.104}{0.107}
& \mse{21.849}{0.052}
& \mse{6.352}{0.261}
& \bestmse{5.016}{0.437}
& \secondmse{5.205}{0.425} \\

Qwen2.5-32B
& \mse{20.637}{0.049}
& \mse{12.545}{0.116}
& \mse{25.448}{0.128}
& \mse{21.091}{0.178}
& \mse{3.624}{0.119}
& \bestmse{3.384}{0.155}
& \secondmse{3.409}{0.146} \\

Qwen2.5-7B
& \mse{32.421}{0.064}
& \mse{22.766}{0.092}
& \mse{32.916}{0.088}
& \mse{27.880}{0.083}
& \mse{6.506}{0.079}
& \bestmse{6.019}{0.075}
& \secondmse{6.033}{0.066} \\

Qwen3-8B
& \mse{27.053}{0.127}
& \mse{15.458}{0.163}
& \mse{33.424}{0.292}
& \mse{27.050}{0.285}
& \mse{6.308}{0.346}
& \bestmse{4.583}{0.198}
& \secondmse{4.588}{0.198} \\

Yi-1.5-34B
& \mse{24.879}{0.084}
& \bestmse{4.626}{0.324}
& \mse{28.171}{0.116}
& \mse{20.323}{0.130}
& \mse{12.845}{0.550}
& \secondmse{10.309}{0.379}
& \mse{10.389}{0.372} \\

Yi-1.5-6B
& \mse{38.327}{0.183}
& \mse{24.587}{0.258}
& \mse{52.420}{0.195}
& \mse{29.660}{0.252}
& \mse{13.834}{0.343}
& \bestmse{10.883}{0.129}
& \secondmse{10.945}{0.117} \\

Yi-1.5-9B
& \mse{33.667}{0.062}
& \mse{21.154}{0.237}
& \mse{49.905}{0.144}
& \mse{39.825}{0.207}
& \mse{2.736}{0.143}
& \bestmse{2.278}{0.228}
& \secondmse{2.312}{0.260} \\

\midrule
\textit{Macro avg.}
& 31.928
& 17.963
& 38.257
& 29.651
& 6.654
& 5.066
& 5.109 \\

\bottomrule
\end{tabular}
\end{adjustbox}

\end{table}

\begin{table}[H]
\centering
\footnotesize
\caption{
MathQA ECE (pp; mean $\pm$ SE over five seeds).
Same conventions as Table~\ref{tab:medmcqa}.
}
\label{tab:mathqa}

\setlength{\tabcolsep}{1.8pt}
\renewcommand{\arraystretch}{1.08}

\begin{adjustbox}{center,max width=0.96\linewidth}
\begin{tabular}{@{}lccccccc@{}}
\toprule
Target PoLM
& Vanilla
& CAPE
& Elicit.
& Elic.-Ens.
& DACA
& \SupportCal{} (1)
& \SupportCal{} (2) \\
\midrule

Gemma3-12B
& \mse{37.034}{0.195}
& \mse{12.704}{0.180}
& \mse{67.366}{0.189}
& \mse{53.653}{0.136}
& \mse{3.073}{0.279}
& \bestmse{0.899}{0.067}
& \secondmse{0.928}{0.085} \\

Llama-3-8B
& \mse{29.188}{0.157}
& \bestmse{1.834}{0.192}
& \mse{50.397}{0.214}
& \mse{39.033}{0.249}
& \mse{6.566}{0.182}
& \secondmse{6.128}{0.134}
& \mse{6.137}{0.134} \\

Qwen2.5-14B
& \mse{24.527}{0.161}
& \mse{7.667}{0.187}
& \mse{42.116}{0.189}
& \mse{34.357}{0.203}
& \secondmse{6.104}{0.450}
& \bestmse{6.075}{0.437}
& \mse{6.136}{0.435} \\

Qwen2.5-32B
& \mse{15.611}{0.113}
& \bestmse{2.118}{0.217}
& \mse{36.599}{0.251}
& \mse{30.307}{0.321}
& \mse{7.486}{0.249}
& \secondmse{7.395}{0.244}
& \mse{7.437}{0.248} \\

Qwen2.5-7B
& \mse{27.715}{0.212}
& \mse{10.858}{0.053}
& \mse{49.017}{0.212}
& \mse{40.547}{0.223}
& \mse{4.044}{0.123}
& \secondmse{3.378}{0.212}
& \bestmse{3.359}{0.216} \\

Qwen2.5-Math-7B
& \mse{13.097}{0.204}
& \bestmse{2.589}{0.039}
& \mse{60.188}{0.138}
& \mse{44.426}{0.199}
& \mse{7.441}{0.528}
& \mse{4.744}{0.518}
& \secondmse{4.639}{0.509} \\

Yi-1.5-9B
& \mse{37.297}{0.187}
& \mse{14.919}{0.139}
& \mse{60.013}{0.184}
& \mse{46.607}{0.183}
& \mse{19.814}{0.511}
& \bestmse{11.255}{0.817}
& \secondmse{11.359}{0.812} \\

\midrule
\textit{Macro avg.}
& 26.353
& 7.527
& 52.242
& 41.276
& 7.790
& 5.696
& 5.714 \\

\bottomrule
\end{tabular}
\end{adjustbox}

\end{table}

Supplementary TweetEval Sentiment results show the same model-level consistency: both tested values of $M$ yield lower mean ECE than DACA on all six targets, reducing the macro-average from 16.59 pp to 15.06 and 15.16 pp (Appendix~\ref{app:tweeteval}).

\textbf{Limited sensitivity to $M$.}
Performance changes little across the two tested reference-combination sizes. 
The maximum absolute per-model ECE difference between the two \SupportCal{} configurations is 0.19 pp on MedMCQA and 0.10 pp on MathQA. 
The corresponding macro-average differences are 0.04 and 0.02 pp.

\textbf{Comparison with other label-free baselines.}
\SupportCal{} achieves the lowest macro-average ECE on both main datasets and is either the best or second-best method for every target model. While CAPE performs strongly on a subset of MathQA targets, \SupportCal{} shows more consistent performance across datasets and model families.

\section{Limitations and Scope}
\label{sec:limitations}

\SupportCal{} uses additional pretrained references only during calibration. This introduces a one-time computational overhead from forward inference to obtain the reference distributions, but requires no additional model training or parameter updates. After fitting $\hat{\tau}$, deployment is identical to ordinary temperature scaling and requires neither additional reference calls nor trainable parameters.

Our current evaluation focuses on fixed-label tasks with a finite output space, including multiple-choice question answering and fixed-label sentiment classification. The present formulation also assumes access to a corresponding own-base PLM and a compatible pool of pretrained references. Extending the framework to settings without a natural own-base reference, or to free-form generation tasks that require a different notion of shared prediction support, remains future work.
\section{Conclusion}
\label{sec:conclusion}

We introduced \SupportCal{}, a label-free calibration method that preserves agreement examples while assigning support-aware weights within the disagreement region.
Motivated by the non-monotonic effect observed under controlled disagreement reintroduction, \SupportCal{} combines the own-base PLM's relative support with corroboration from selected pretrained references and uses the resulting weights to fit a single scalar temperature.
Across MedMCQA and MathQA, \SupportCal{} yields lower macro-average ECE than agreement-only calibration, with the same overall trend on supplementary TweetEval Sentiment. 
Our boundary characterization further links the weighted objective to whether the optimal temperature remains finite.
Overall, these results show that support-aware weighting can extract useful calibration information from disagreement examples while retaining the simplicity and deployment cost of scalar temperature scaling.

\FloatBarrier
\bibliographystyle{plainnat}
\bibliography{references}

\appendix
\section{Own-base disagreement rate by target model}
\label{app:disagreement}
\begin{figure}[H]
\centering
\begin{minipage}[t]{0.49\textwidth}
\vspace{0pt}
\centering
\includegraphics[width=\linewidth]{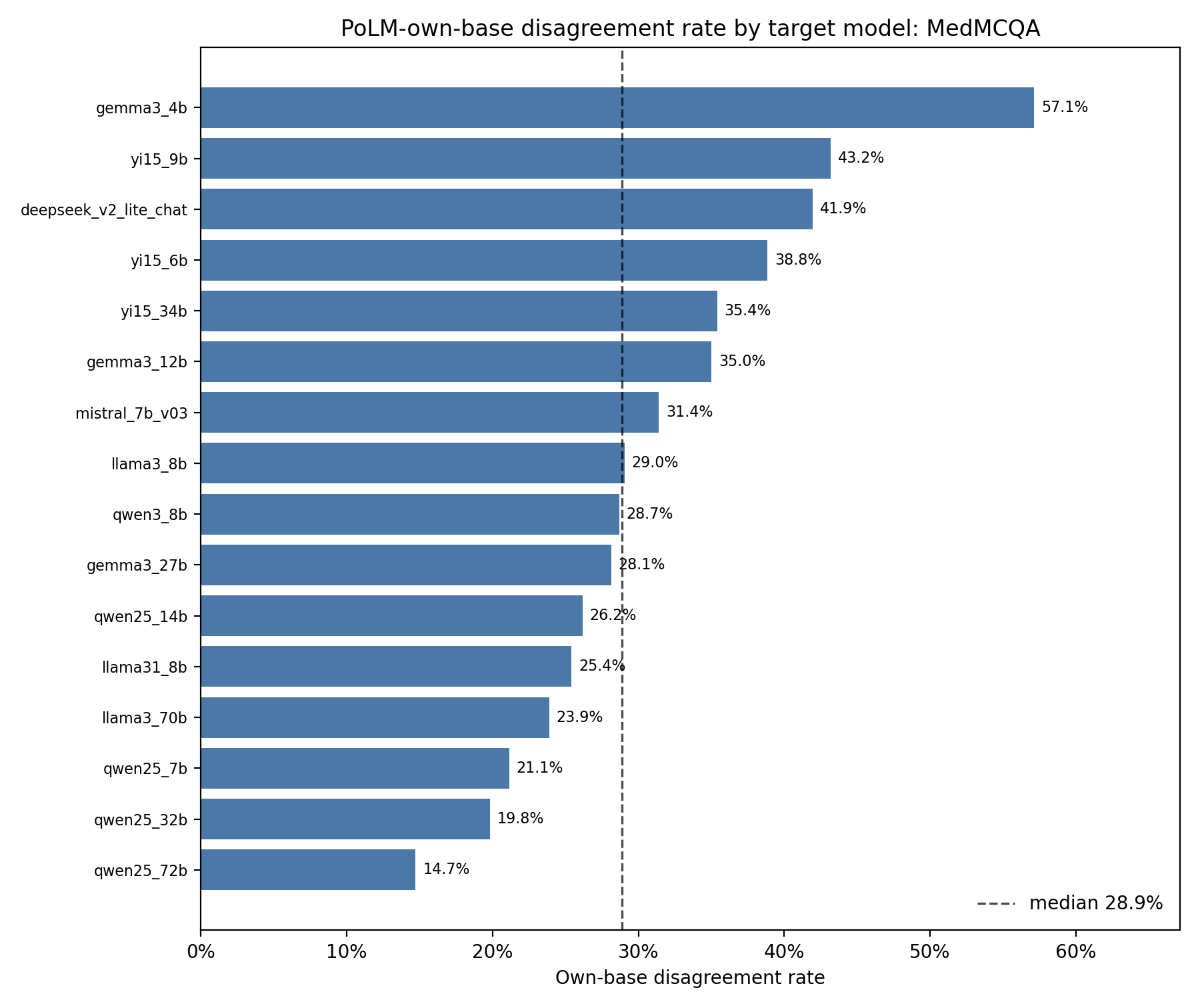}
\small (a) MedMCQA, median 28.9\%.
\end{minipage}\hfill
\begin{minipage}[t]{0.49\textwidth}
\vspace{0pt}
\centering
\includegraphics[width=\linewidth]{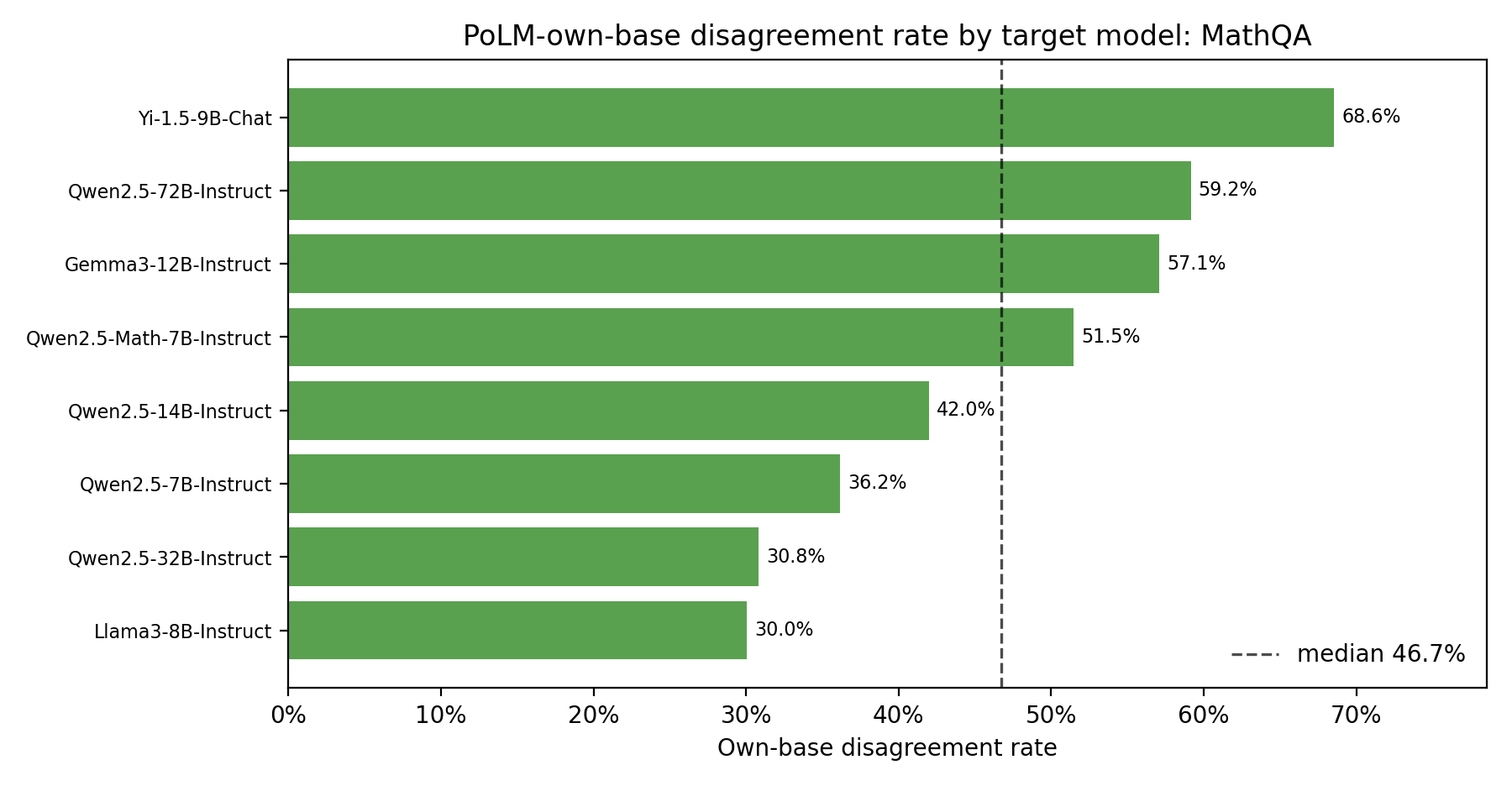}
\small (b) MathQA, median 46.7\%.
\end{minipage}
\caption{Own-base PLM/PoLM disagreement rates across the broader diagnostic
target pool.}
\label{fig:disagreement-rate}
\end{figure}

\section{TweetEval Sentiment: supplementary validation}
\label{app:tweeteval}

We additionally evaluate \SupportCal{} on TweetEval Sentiment
\citep{barbieri2020tweeteval}, a three-class non-QA classification task, using six
target PoLMs under the same protocol as the main experiments. Table~\ref{tab:tweeteval}
reports per-model ECE against DACA for $M\in\{2,4\}$.

\begin{table}[H]
\centering
\small
\caption{
TweetEval Sentiment: mean ECE $\pm$ SE over five seeds.
Bold and underline denote the best and second-best values among the methods shown;
lower is better.
}
\label{tab:tweeteval}

\setlength{\tabcolsep}{6pt}
\renewcommand{\arraystretch}{1.08}

\begin{tabular}{@{}lccc@{}}
\toprule
Target PoLM
& DACA
& \SupportCal{} ($M=2$)
& \SupportCal{} ($M=4$) \\
\midrule

Llama-3-8B
& $13.7598\pm0.3353$
& $\underline{11.4970\pm0.3579}$
& $\mathbf{11.4871\pm0.3665}$ \\

Llama-3.1-8B
& $15.5236\pm0.5474$
& $\mathbf{15.1181\pm0.1213}$
& $\underline{15.1251\pm0.1222}$ \\

Qwen2.5-7B
& $17.7535\pm0.4926$
& $\mathbf{16.4443\pm0.2955}$
& $\underline{16.4889\pm0.2881}$ \\

Qwen2.5-14B
& $21.6643\pm0.1817$
& $\mathbf{19.6526\pm0.3147}$
& $\underline{20.0094\pm0.3145}$ \\

Qwen2.5-32B
& $14.9633\pm0.4463$
& $\mathbf{14.0326\pm0.3799}$
& $\underline{14.2642\pm0.3669}$ \\

Qwen3-8B
& $15.8799\pm0.4372$
& $\underline{13.6205\pm0.4459}$
& $\mathbf{13.6063\pm0.4531}$ \\

\bottomrule
\end{tabular}
\end{table}

\section{Additional setup details}
\label{app:setup}

\textbf{Size-compatible reference pools.}
The model names below denote pretrained/base checkpoints. The small bucket contains 
Llama-3-8B, Llama-3.1-8B, Qwen2.5-7B, Qwen3-8B, Gemma3-4B, Yi-1.5-6B, Yi-1.5-9B, Mistral-7B-v0.3
\citep{mistralai2024mistral7bv03}, and DeepSeek-V2-Lite
\citep{deepseekai2024deepseekv2}. 
The medium bucket contains Qwen2.5-14B, Qwen2.5-32B, Gemma3-12B, Gemma3-27B, and Yi-1.5-34B.
For each dataset and target model, the candidate pool $R$ contains the available same-bucket pretrained references, excluding the target's own-base PLM. 
For each fixed $M$, we enumerate every eligible
$S\subseteq R$ with $|S|=M$ and select the combination with the largest Disagreement-Support score (Eq.~\ref{eq:selector}).
\textbf{Tested combination sizes.}
MedMCQA uses $M\in\{2,4\}$, MathQA uses $M\in\{1,2\}$, and TweetEval Sentiment uses $M\in\{2,4\}$.

\section{Role of cross-reference corroboration}
\label{app:relative_only_failure}

The own-base relative-support score provides an informative sample-level signal, while cross-reference corroboration supplies complementary evidence. Tables~\ref{tab:relative_only_medmcqa} and~\ref{tab:relative_only_mathqa} report representative targets where relative-only weighting is less effective than the full \SupportCal{} construction. The MedMCQA comparison includes DeepSeek-V2-Lite from the broader diagnostic target pool in Appendix~\ref{app:disagreement}.

\begin{table}[H]
\centering
\small
\caption{
Representative MedMCQA relative-support ablation. Mean ECE $\pm$ SE over five seeds;
lower is better.
}
\label{tab:relative_only_medmcqa}

\setlength{\tabcolsep}{6pt}
\renewcommand{\arraystretch}{1.08}

\begin{adjustbox}{center,max width=\linewidth}
\begin{tabular}{@{}lcccc@{}}
\toprule
Target PoLM
& DACA
& Relative-only
& \SupportCal{} ($M=2$)
& \SupportCal{} ($M=4$) \\
\midrule

DeepSeek-V2-Lite
& $1.5632\pm0.2065$
& $3.0482\pm0.2445$
& $1.5589\pm0.2522$
& $\mathbf{1.4886\pm0.2678}$ \\

Yi-1.5-9B
& $2.7360\pm0.1426$
& $3.7620\pm0.3241$
& $\mathbf{2.2775\pm0.2282}$
& $2.3115\pm0.2597$ \\

\bottomrule
\end{tabular}
\end{adjustbox}
\end{table}

\begin{table}[H]
\centering
\small
\caption{
Representative MathQA relative-support ablation. Mean ECE $\pm$ SE over five seeds;
lower is better.
}
\label{tab:relative_only_mathqa}

\setlength{\tabcolsep}{6pt}
\renewcommand{\arraystretch}{1.08}

\begin{adjustbox}{center,max width=\linewidth}
\begin{tabular}{@{}lcccc@{}}
\toprule
Target PoLM
& DACA
& Relative-only
& \SupportCal{} ($M=1$)
& \SupportCal{} ($M=2$) \\
\midrule

Llama-3-8B
& $6.5657\pm0.1818$
& $6.9351\pm0.1898$
& $\mathbf{6.1277\pm0.1337}$
& $6.1368\pm0.1343$ \\

Qwen2.5-14B
& $6.1044\pm0.4504$
& $8.4290\pm0.1791$
& $\mathbf{6.0750\pm0.4377}$
& $6.1363\pm0.4349$ \\

Qwen2.5-32B
& $7.4862\pm0.2489$
& $9.2408\pm0.3519$
& $\mathbf{7.3953\pm0.2435}$
& $7.4374\pm0.2489$ \\

\bottomrule
\end{tabular}
\end{adjustbox}
\end{table}

\section{Boundary characterization and empirical weight--margin relationship}
\label{app:boundary}

This appendix proves Proposition 1 (Section~\ref{sec:boundary}). It then reports an
empirical check of how \SupportCal{} weights vary with the boundary margin identified by
the derivation.

\subsection{Weighted KL objective in inverse-temperature coordinates}
\label{app:boundary-inverse}
Starting from the temperature-fitting objective (Eq.~\ref{eq:tempobj}), whose original
domain is $\tau>0$, substitute $\beta=1/\tau$ and extend the domain to
$\beta\in[0,\infty)$: for $\beta>0$ this is equivalent to $\tau=1/\beta>0$, while the
boundary point $\beta=0$ is not itself in the original domain and corresponds to the
limit $\tau\to\infty$. Write $q_{n,\beta}=\mathrm{softmax}(\beta z_n)$. Writing
$W=\sum_n w_n$, the weights in Eq.~\eqref{eq:weights} are strictly positive for any
nonempty calibration split: agreement examples have $w_n=1$, and disagreement examples
have $u_n^{(o)}>0$ and $c_n^\star>0$ because finite-logit softmax distributions assign
positive probability to every class. Thus $W>0$, and the objective becomes
$L(\beta)=\frac{1}{W}\sum_n w_n \ell_n(\beta)$ with
$\ell_n(\beta)=D_{\mathrm{KL}}(p_n^{(o)}\,\|\,q_{n,\beta})$. Expanding the KL divergence,
\begin{equation}
\ell_n(\beta) = C_n + \log\sum_{k=1}^K e^{\beta z_{n,k}} - \beta\,\mathbb{E}_{p_n^{(o)}}[z_n],
\label{eq:ell}
\end{equation}
where $C_n=\sum_k p^{(o)}_{n,k}\log p^{(o)}_{n,k}$ is the (negative) entropy of
$p_n^{(o)}$ and does not depend on $\beta$, and
$\mathbb{E}_{p_n^{(o)}}[z_n]=\sum_k p^{(o)}_{n,k} z_{n,k}$.

\subsection{First derivative}
Differentiating the log-sum-exp term in Eq.~\eqref{eq:ell},
$\frac{d}{d\beta}\log\sum_k e^{\beta z_{n,k}} = \sum_k z_{n,k}\,q_{n,\beta,k} =
\mathbb{E}_{q_{n,\beta}}[z_n]$, so
\begin{gather}
\ell_n'(\beta) = \mathbb{E}_{q_{n,\beta}}[z_n] - \mathbb{E}_{p_n^{(o)}}[z_n],
\label{eq:firstderiv}
\\
L'(\beta) = \frac{1}{W}\sum_n w_n\,\ell_n'(\beta).
\notag
\end{gather}
The derivative compares the average target-PoLM logit under the current scaled
distribution $q_{n,\beta}$ against the same average under the own-base reference
$p_n^{(o)}$.

\subsection{Second derivative and convexity}
\label{app:boundary-convexity}
Since $\frac{d}{d\beta}q_{n,\beta,k}=q_{n,\beta,k}\big(z_{n,k}-\mathbb{E}_{q_{n,\beta}}[z_n]\big)$,
differentiating Eq.~\eqref{eq:firstderiv} gives
\begin{gather}
\ell_n''(\beta) = \mathrm{Var}_{q_{n,\beta}}(z_n) \ge 0,
\label{eq:secondderiv}
\\
L''(\beta) = \frac{1}{W}\sum_n w_n\,\mathrm{Var}_{q_{n,\beta}}(z_n) \ge 0,
\notag
\end{gather}
Since $w_n>0$ for all $n$, Eq.~\eqref{eq:secondderiv} shows that $L$ is convex on $[0,\infty)$. 
For every finite $\beta$, $q_{n,\beta}$ has full support. Consequently,
\[
\mathrm{Var}_{q_{n,\beta}}(z_n)>0
\]
whenever $z_n$ is nonconstant. If at least one calibration example has a nonconstant target-logit vector, its positive weight therefore gives $L''(\beta)>0$ for every finite $\beta$, and $L$ is strictly convex on $[0,\infty)$.

\subsection{\texorpdfstring{Boundary at $\beta=0$}{Boundary at beta = 0}}

At $\beta=0$, the scaled distribution is uniform,
$q_{n,0,k}=1/K$, so
$\mathbb{E}_{q_{n,0}}[z_n]
=\bar z_n:=K^{-1}\sum_{k=1}^{K}z_{n,k}$.
Define the per-example \emph{boundary margin}
\begin{equation}
m_n
:=
\mathbb{E}_{p_n^{(o)}}[z_n]
-
\bar z_n .
\label{eq:margin}
\end{equation}

Substituting into Eq.~\eqref{eq:firstderiv} gives
$\ell_n'(0)=-m_n$, so $m_n$ captures the signed per-example
contribution to the objective's boundary derivative.

Define the weighted boundary margin
$M_w:=W^{-1}\sum_n w_n m_n$. Averaging the per-example derivatives
then gives
\begin{equation}
L'(0)
=
\frac{1}{W}\sum_n w_n\ell_n'(0)
=
-\frac{1}{W}\sum_n w_n m_n
=
-M_w.
\label{eq:boundaryderiv}
\end{equation}

\subsection{Finite versus infinite temperature}
In our setting, every own-base distribution $p_n^{(o)}$ has full support because it is a finite-logit softmax output. The evaluated model outputs are nondegenerate, so at least one positive-weight calibration example has nonconstant target logits. 
Together with $W>0$, Section~\ref{app:boundary-convexity} gives strict convexity on $[0,\infty)$ and a strictly increasing
derivative $L'$.

\emph{Case $M_w\le0$.} Eq.~\eqref{eq:boundaryderiv} gives $L'(0)\ge0$. 
Since convexity makes $L'$ nondecreasing, $L'(\beta)\ge0$ for all $\beta\ge0$; hence $L$ is minimized on the extended domain at $\beta^\star=0$. Because $\beta=0$ lies outside the original domain $\tau>0$ (Section~\ref{app:boundary-inverse}), the original objective has no finite minimizing temperature and its infimum is approached only as $\tau\to\infty$.

\emph{Case $M_w>0$.} Eq.~\eqref{eq:boundaryderiv} gives $L'(0)<0$. 
As $\beta\to\infty$, $q_{n,\beta}$ concentrates on the maximizers of $z_n$, so
$\lim_{\beta\to\infty}\ell_n'(\beta)=\max_k z_{n,k}-\mathbb{E}_{p_n^{(o)}}[z_n]$. 
Because $p_n^{(o)}$ has full support, this limit is positive for any nonconstant $z_n$:
$\mathbb{E}_{p_n^{(o)}}[z_n]<\max_k z_{n,k}$. 
The positive-weight nonconstant example therefore makes $\lim_{\beta\to\infty}L'(\beta)>0$. 
Continuity and strict monotonicity of $L'$ give exactly one zero, at $\beta^\star\in(0,\infty)$, and strict convexity makes it
the unique global minimizer. 
Equivalently, $\tau^\star=1/\beta^\star$ is finite and unique.

The boundary condition is therefore determined by $M_w$: $M_w>0$ yields a unique finite-temperature optimum, while $M_w\le0$ places the optimum at the $\tau\to\infty$ boundary. 
\SupportCal{} controls each disagreement example's contribution through its weight.

\subsection{Empirical weight--margin relationship}
\begin{figure}[H]
\centering
\includegraphics[width=\linewidth]{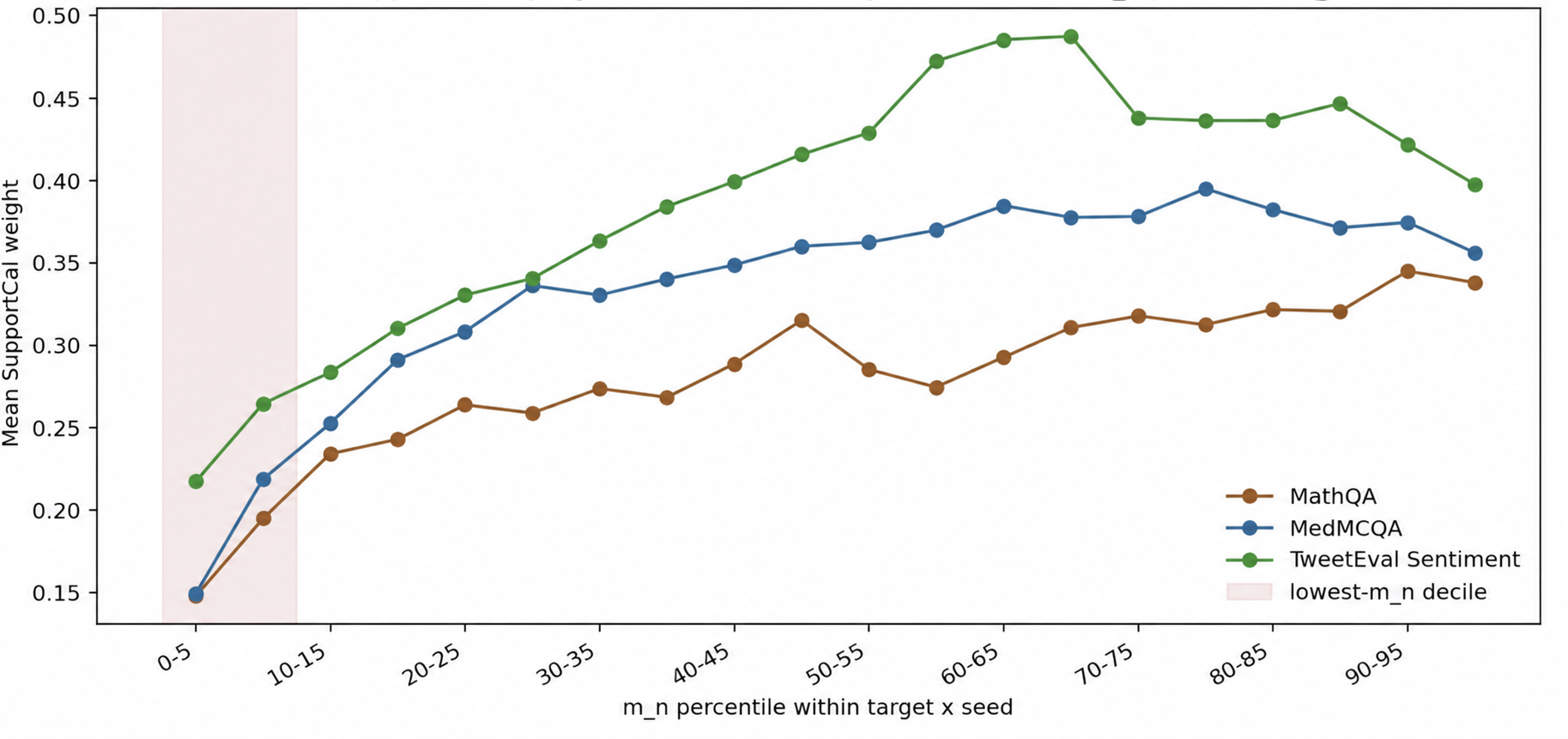}
\caption{Mean \SupportCal{} sample weight by $m_n$ percentile bin, on the disagreement
region of MedMCQA, MathQA, and TweetEval Sentiment. The shaded band marks the
lowest-$m_n$ decile.}
\label{fig:weightmargin}
\end{figure}

Figure~\ref{fig:weightmargin} shows a broad association between the boundary margin and the disagreement weight assigned by \SupportCal{}. Samples in the lowest $m_n$ percentiles receive the smallest average weights, and the mean weight generally increases with $m_n$. 
The sign-based summary in Table~\ref{tab:signsplit-weights} shows the same pattern more directly: across all three datasets, disagreement examples with $m_n<0$ receive lower mean and median weights than those with $m_n>0$. This pattern is consistent with the weighting rule attenuating examples that exert pressure toward the $\beta=0$ ($\tau\to\infty$) boundary identified by the analysis above.

\begin{table}[H]
\centering
\small
\caption{
\SupportCal{} disagreement weights by boundary-margin sign (mean / median).
}
\label{tab:signsplit-weights}

\setlength{\tabcolsep}{10pt}
\renewcommand{\arraystretch}{1.12}

\begin{tabular}{@{}lcc@{}}
\toprule
Dataset & $m_n<0$ & $m_n>0$ \\
\midrule
MathQA
& 0.187 / 0.120
& 0.316 / 0.249 \\

MedMCQA
& 0.244 / 0.199
& 0.372 / 0.338 \\

TweetEval Sentiment
& 0.340 / 0.294
& 0.431 / 0.404 \\

\bottomrule
\end{tabular}
\end{table}

\end{document}